\documentclass[10pt,twocolumn,letterpaper]{article}

\usepackage{cvpr}              

\usepackage{graphicx}
\usepackage{amsmath}
\usepackage{amssymb}
\usepackage{booktabs}

\usepackage{multirow}
\usepackage{tabularx,verbatim}
\usepackage{xspace}
\usepackage{algorithm}
\usepackage{algorithmic}
\usepackage{setspace}
\usepackage{comment}
\usepackage{bbm}
\usepackage{enumitem}
\usepackage{colortbl}
\usepackage{color, xcolor} 
\usepackage[T1]{fontenc}
\usepackage{bbm}
\usepackage{listings}
\usepackage{subcaption}
\usepackage{pifont}
\usepackage{fontawesome}
\usepackage[symbol]{footmisc}

\definecolor{remark}{rgb}{1,.5,0} 
\definecolor{citecolor}{rgb}{0,0.443,0.737} 
\definecolor{linkcolor}{rgb}{0.956,0.298,0.235} 
\definecolor{cyan}{rgb}{0.831,0.901,0.945}
\definecolor{antiquewhite}{rgb}{0.98, 0.92, 0.84}
\definecolor{myred}{RGB}{176, 36, 24}
\definecolor{myblue}{RGB}{79,113,190}

\usepackage{amsmath}

\definecolor{cvprblue}{rgb}{0.21,0.49,0.74}
\usepackage[pagebackref,breaklinks,colorlinks,citecolor=cvprblue]{hyperref}

\def\paperID{1252} 
\def\confName{CVPR}
\def\confYear{2024}

\newcommand{\ours}{Make-A-Protagonist\xspace}

\title{Make-A-Protagonist: Generic Video Editing with Visual and Textual Clues}

\author{%
Yuyang Zhao$^1$,
Enze Xie$^2$\textsuperscript{\faEnvelopeO}, 
Lanqing Hong$^1$,
Zhenguo Li$^3$,
Gim Hee Lee$^1$ \\
{$^1$ National University of Singapore}  \\
{$^2$ The University of Hong Kong }\\ 
{$^3$ The Hong Kong University of Science and Technology}\\ 
\textbf{\url{https://make-a-protagonist.github.io}}
\vspace{-.1in}
}

\begin{document}

\twocolumn[{
      \vspace{-.5em}
      \maketitle
      \begin{center}
        \centering
        \vspace{-0.1in}
        \includegraphics[width=\linewidth]{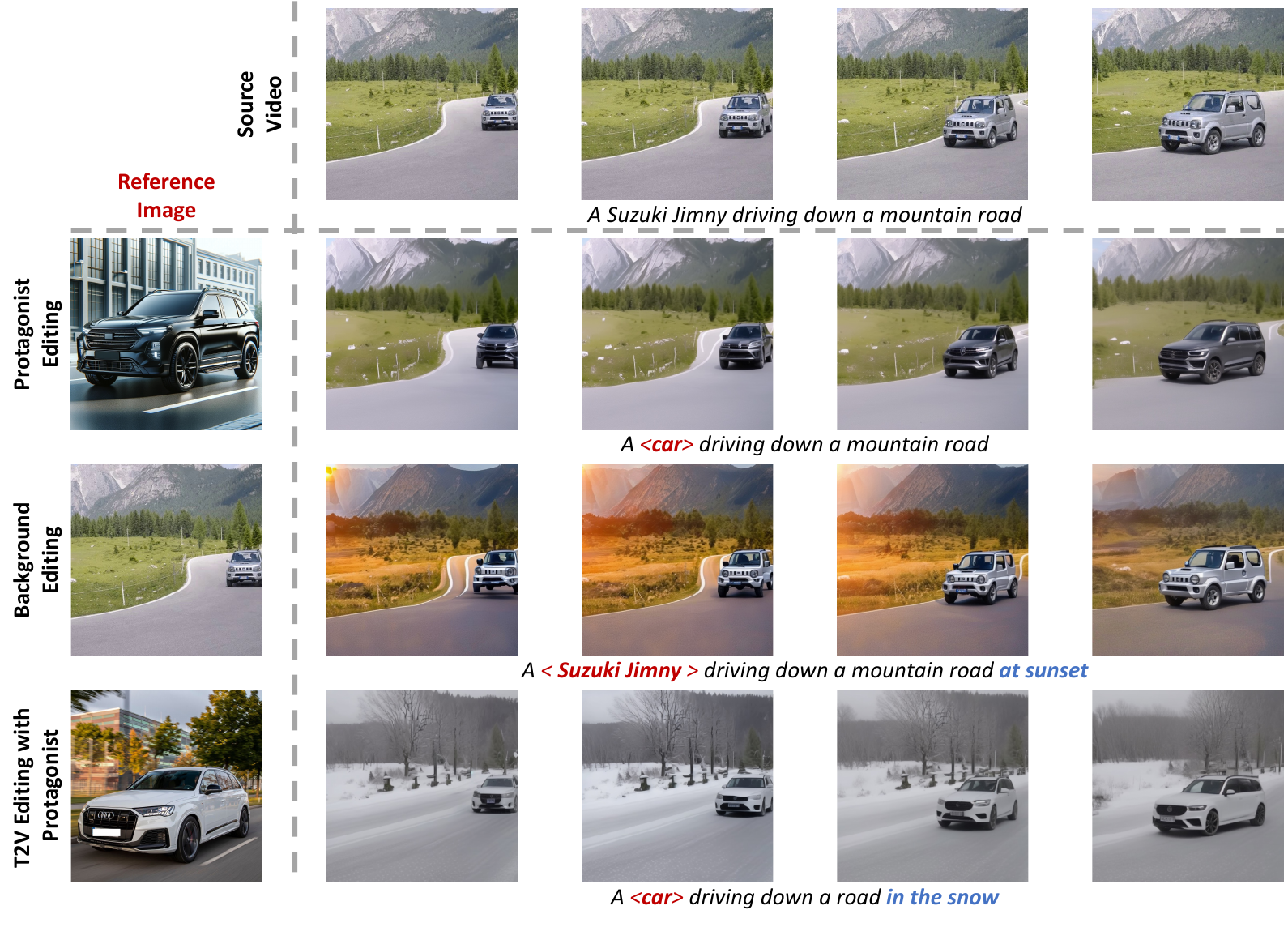}
        \vspace{-.3in}
        \captionof{figure}{Three applications of \ours. \textcolor{myred}{<\textit{car}>} is the protagonist in the video. \textcolor{myblue}{\textbf{\textit{Bold blue}}} denotes the textual description. 
        }
        \label{fig:teaser}
      \end{center}
    }]
    
\begin{abstract}
The text-driven image and video diffusion models have achieved unprecedented success in generating realistic and diverse content. Recently, the editing and variation of existing images and videos in diffusion-based generative models have garnered significant attention.
However, previous works are limited to editing content with text or providing coarse personalization using a single visual clue, rendering them unsuitable for indescribable content that requires fine-grained and detailed control.
In this regard, we propose a generic video editing framework called \ours, which utilizes textual and visual clues to edit videos with the goal of empowering individuals to become the protagonists.
Specifically, we design a visual-textual-based video generation model coupled with a mask-guided fusion method to integrate source video, target visual and textual clues.
Extensive results demonstrate the versatile and remarkable editing capabilities of \ours.

\end{abstract}

\section{Introduction}
\label{sec:intro}

\begin{center}
\begin{quote}
\textit{``The protagonist sets the scene for the entire story.''} 
--- Ben Okri~\cite{quote}
\end{quote}
\end{center}

Diffusion-based generative models have demonstrated remarkable success in generating photorealistic and diverse images~\cite{stablediffusion,saharia2022photorealistic,unclip} and videos~\cite{makeavideo,zhou2022magicvideo,wu2022tune} conditioned on text. However, the generation process, while more diverse, lacks controllability when relying solely on text descriptions.
To generate the desired content, researchers explore two approaches to modify text-conditioned generation: incorporating external control signals~\cite{zhang2023adding, mou2023t2i, lhhuang2023composer, gen1} and editing existing content with textual information~\cite{hertz2022prompt, qi2023fatezero,liu2023videop2p,meng2021sdedit}. 
However, both approaches rely on text descriptions to convey the desired content, which raises an important question: \textit{what if the content cannot be accurately described through text?}
As an illustration, we consider the source video shown in Figure~\ref{fig:teaser}.
Previous studies~\cite{qi2023fatezero,liu2023videop2p} are capable of replacing the ``Suzuki Jimny'' with a recognizable brand, such as ``Porsche'' or ``Mercedes-Benz'' since these options are identifiable by the text model. Nevertheless, they are unable to substitute it with an unnamed car. For example, the reference image in 2rd row of Fig.~\ref{fig:teaser} is generated by a text-to-image model with the text prompt ``a black car'', so it does not have a corresponding name that can be precisely recognized by the language model.

In contrast, people have the strong motivation to personalize their own content and to change the \textit{protagonist}, which may not be accurately described through text.
This motivation has served as inspiration for the development of personalized models~\cite{ruiz2022dreambooth,molad2023dreamix} and image/video variation techniques~\cite{unclip,gen1}. 
Nevertheless, personalized models necessitate fine-tuning for each specific target using multiple target images and exhibit sensitivity to hyperparameters and training configuration.
Furthermore, image and video variation models tend to exhibit bias toward the background of the reference images.
Additionally, both types of generation models are typically limited to generating content based on a single visual reference clue, which proves ineffective when multiple protagonists need to be modified.
These limitations motivate us to design a framework for generic video editing with text and an arbitrary number of reference images.

One recent work~\cite{wu2022tune} can be applied to generic video editing by combining personalized modeling~\cite{ruiz2022dreambooth} with one-shot video editing. Specifically, this work first fine-tunes a text-to-image generation (T2I) model with the reference image. After that, it inflates the T2I model with temporal layers and optimizes it on the source video. However, the image-video tuning process has a severe limitation: the two models should be retrained when changing the reference image. In addition, it also inherits the issues of personalized modeling, sensitive to hyperparameters.
To this end, this paper disentangles the personally edited content (protagonists) from the source video to realize end-to-end one-stage generic video editing.
To achieve generic video editing, the framework necessitates the ability of leveraging visual clues, leveraging textual clues and maintaining source motion.
Therefore, we build our framework based on Stable UnCLIP~\footnote{https://huggingface.co/stabilityai/stable-diffusion-2-1-unclip-small}, which takes CLIP image and text embeddings as conditions. The image embedding is directly added to the model features while text embedding is utilized via cross-attention. 
Following~\cite{makeavideo,zhou2022magicvideo,wu2022tune}, we inflate the T2I model (Stable UnCLIP) to one-shot text-to-video generation (T2V) model by including additional temporal modules.
After tuning the T2V model with the source video, we propose a novel mask-guided fusion to incorporate the visual clue, textual clue and source motion with the mask of the protagonist of the source video. The mask is obtained from pre-trained models~\cite{groundingdino,sam,cheng2022xmem}, eliminating the need for data annotation.
With the diffusion-based T2V model and effective mask-guided fusion method, our framework (\ours) achieves commendable performance in both conventional and novel video editing tasks, including \textit{protagonist editing while keeping the background}, \textit{background editing of the source content}, and \textit{text-to-video editing with the protagonist}. 
Our contributions can be summarized as:
\begin{itemize}
    \item We present the first end-to-end framework for generic video editing with both visual and textual clues.
    \item We design a visual-textual-based video generation model and a novel mask-guided fusion to incorporate the visual clue, textual clue and source motion, realizing strong video editing performance.
    \item Extensive results demonstrate the versatile applications of \ours and the superiority over previous video editing works.
\end{itemize}

\section{Related Work}
\label{sec:related-work}

\begin{figure*}[t]
    \centering
    \includegraphics[width=.99\textwidth]{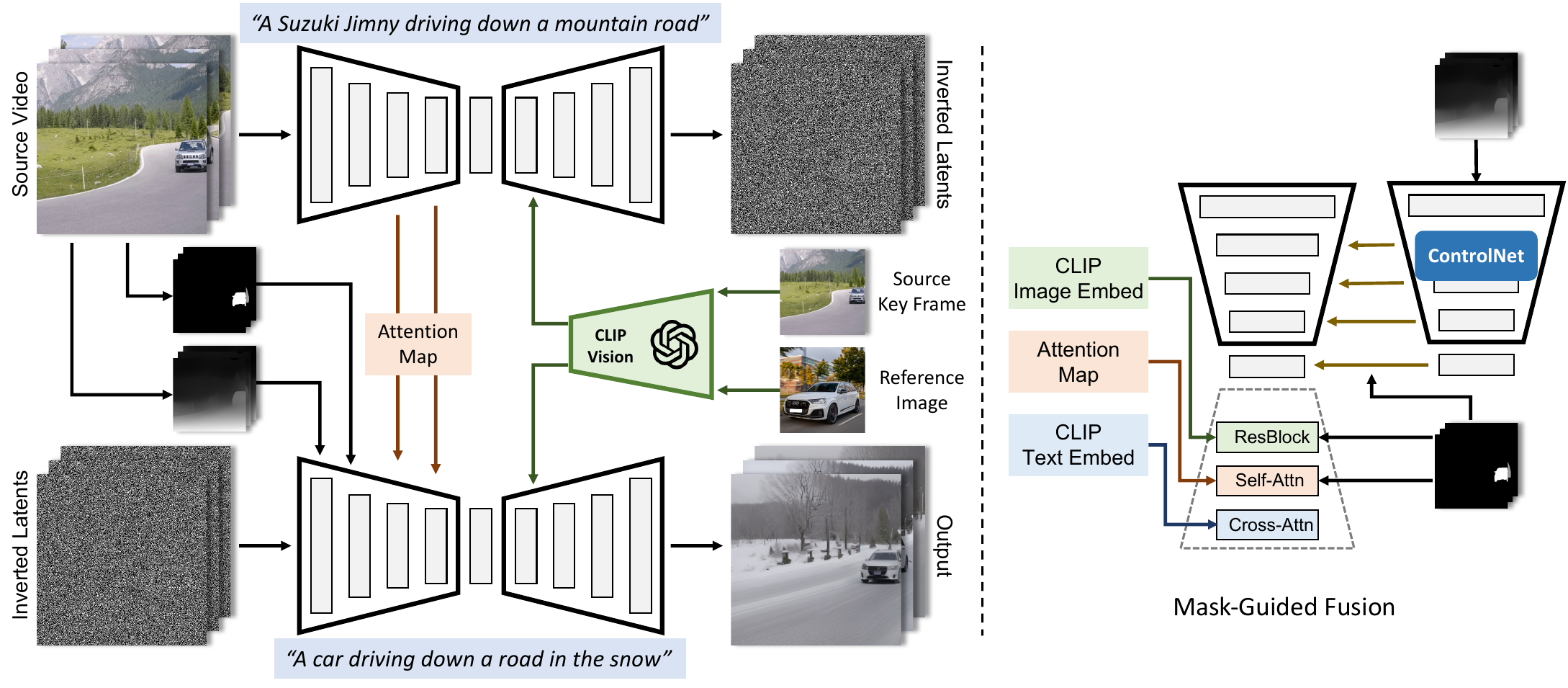}
    \vspace{-.05in}
    \caption{The overall inference framework of \ours.  
    Masks and control signals are extracted from the source video. Source video, visual and textual clues are fused in the video generation model with the mask-guided fusion to enable generic video editing.
    }
    \label{fig:framework}
\end{figure*}

\noindent\textbf{Visual Content Generation.}
Content generation has made remarkable advancements with powerful generative models~\cite{van2017neural, yu2022scaling,kang2023scaling}. Recently, equipped with vast and diverse image-text pairs from the internet, diffusion-based generative models~\cite{unclip,stablediffusion,saharia2022photorealistic} have outperformed GAN-based methods in the text-to-image generation (T2I). In view of the impressive performance in T2I, text-to-video generation (T2V)~\cite{zhou2022magicvideo,makeavideo,ho2022imagen, luo2023decomposed, hong2022cogvideo} has attracted much attention. T2V models often utilize pre-trained T2I models to leverage the abundant image-text resources. Additionally, temporal modules~\cite{makeavideo,zhou2022magicvideo} are introduced to facilitate video representation learning. Furthermore, temporal consistency is also established in latent space to enhance video generation~\cite{luo2023decomposed,khachatryan2023text2video}.

\noindent\textbf{Visual Content Editing and Variation with Text.}
An alternative direction for content generation is the editing of existing images~\cite{hertz2022prompt,meng2021sdedit,tumanyan2022plug,brooks2022instructpix2pix} and videos~\cite{liu2023videop2p,qi2023fatezero,shin2023edit,bar2022text2live, wu2022tune} using text, instead of uncontrolled generation solely based on text descriptions.
SDEdit~\cite{meng2021sdedit} applies noise to the image and recovers the image for editing purposes.
Prompt-to-prompt~\cite{hertz2022prompt} and Plug-and-Play~\cite{tumanyan2022plug} modify the cross-attention map by changing the text description.
For video editing, Text2Live~\cite{bar2022text2live} divides the video into layers and edits each layer separately using a text description. Tune-A-Video~\cite{wu2022tune} inflates and fine-tunes the T2I model on a single video and generates new videos of similar motion. 
Video-P2P~\cite{liu2023videop2p} and FateZero~\cite{qi2023fatezero} extend the Prompt-to-prompt into video level. TokenFlow~\cite{tokenflow2023} edits several key frames with prompt-to-prompt and then propagates them to the whole video via feature matching.
However, previous methods solely focus on content editing through text, making them unsuitable for indescribable content.

\noindent\textbf{Visual Content Variation and Personalization.}
To address the indescribable content, DALL-E~2~\cite{unclip} and Gen-1~\cite{gen1} perform image and video variation using CLIP~\cite{clip} image embedding. However, since the CLIP vision model extracts information from the whole image, background information is inevitably incorporated into the variation. 
Personalized models~\cite{molad2023dreamix,ruiz2022dreambooth,textualinversion} present an alternative approach for handling indescribable content. However, DreamBooth~\cite{ruiz2022dreambooth} and DreamMix~\cite{molad2023dreamix} necessitate multiple images to fine-tune a T2I and T2V model for concept learning, and the fine-tuning process is sensitive to training configuration.
Considering the limitations of previous works, this paper presents a framework for generic video editing with both image and text descriptions.

\section{Make-A-Protagonist}
\subsection{Overview}
To realize generic video editing with both visual and textual clues, we introduce a video generation model that can take both CLIP image embedding and CLIP text embedding as conditions (Sec.~\ref{sec:videosd}). In addition, since the CLIP image embedding is directly added to the features, it lacks spatial control compared with the cross-attention used by text embedding. Therefore, we propose a mask-guided fusion technique, leveraging the mask of the protagonist in the source video to accurately control the spatial location during inference (Sec.~\ref{sec:mask-denoise}).  Furthermore, we introduce attention fusion and ControlNet to precisely control the semantic and spatial information of the background and protagonist, respectively.
Equipped with the video generation model and mask-guided fusion, our framework achieves three applications (Sec.~\ref{sec:applications}).
The overall framework is depicted in Fig.~\ref{fig:framework}.

\begin{figure*}[t]
    \centering
    \includegraphics[width=.9\textwidth]{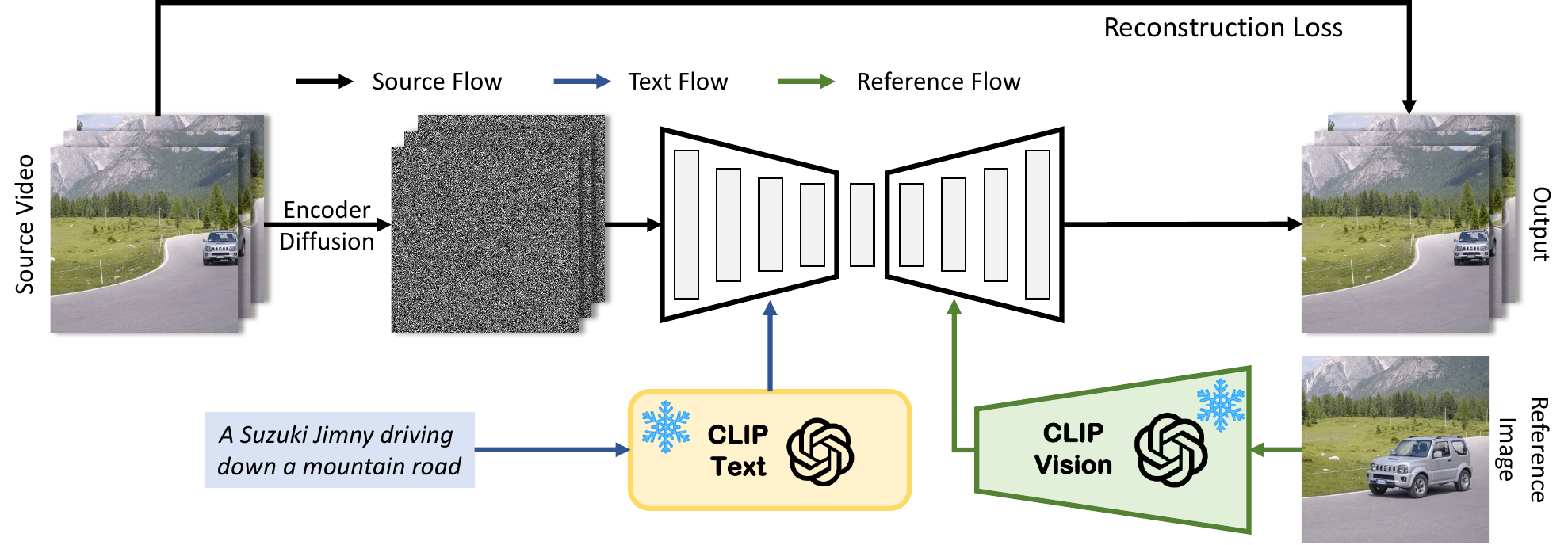}
    \vspace{-.05in}
    \caption{Training process of visual-textual-based video generation model. The model is trained with the video caption and randomly sampled reference frame.}
    \label{fig:train-process}
\end{figure*}

\subsection{Visual-Textual-based Video Generation Model}
\label{sec:videosd}

\textbf{Latent Diffusion Models (LDMs) with Image Embedding.}
LDMs consist of two main components: an autoencoder that encodes an RGB image $x$ into a latent $z=\mathcal{E}(x)$ with the encoder $\mathcal{E}$, and a decoder $\mathcal{D}$ that reconstructs the image $x\approx \mathcal{D}(z)$ from $z$.
In the latent space, a U-Net~\cite{unet} $\varepsilon_\theta$ containing residual blocks and self-/cross-attention is used to remove the noise with the objective:
\begin{equation}
\label{eq:train-objective}
\begin{aligned} 
\min _\theta E_{z_0, \varepsilon, t }
\left\| \sqrt{\alpha_t}\cdot \varepsilon - \sqrt{1-\alpha_t} \cdot z_0 -\varepsilon_\theta\left(z_t, t, \mathcal{C}, \mathcal{I}\right)
\right\|_2^2,
\end{aligned}
\end{equation}
where $\mathcal{C}$ and $\mathcal{I}$ denote the embedding of the text and image, respectively. Noise $\varepsilon$ is added to $z_0$ according to step $t$ to obtain $z_t$. 
Note that instead of directly predicting the noise $\varepsilon$~\cite{ho2020denoising}, following~\cite{gen1,ho2022imagen}, we use $v$-parameterization~\cite{vprediction} to improve color consistency of videos.
Regarding text and image conditions, the text condition serves as the key and value for cross-attention while the image condition is directly added to features of residual blocks, which aligns with \cite{stablediffusion} and \cite{unclip}, respectively.

\noindent\textbf{DDIM Inversion.}
Deterministic DDIM sampling is used to generate each frame from a latent noise in $T$ denoising steps:
\begin{equation}
\label{eq:ddim}
\begin{aligned}
    z_{t-1} &= \sqrt{\alpha_{t-1}} \left(\sqrt{\alpha_{t}}~ z_t - \sqrt{1 - \alpha_{t}}~ \varepsilon_\theta \right) \\
    &+ \sqrt{1 - \alpha_{t-1}} \left(\sqrt{\alpha_{t}}~ \varepsilon_\theta + \sqrt{1 - \alpha_{t}}~ z_t \right),
\end{aligned}
\end{equation}
where $t : 1 \rightarrow T$ denotes the timestamp, $\alpha_{t}$ is a parameter for noise scheduling~\cite{song2020denoising}. Same with Eq.~\ref{eq:train-objective}, we use $v$-parameterization for denoising. 

Since \ours focuses on video editing, instead of using random noise during inference, we use DDIM inversion~\cite{song2020denoising} to transform the source video into the initial latent code that requires denoising. This approach enables the preservation of both the motion and structural information presented in the source video, while also enhancing the temporal consistency within the latent space. The DDIM inversion is illustrated as:
\begin{equation}
\begin{aligned}
\label{eq:ddim-inv}
    z_{t+1} &= \sqrt{\alpha_{t+1}} \left(\sqrt{\alpha_{t}}~ z_{t} - \sqrt{1 - \alpha_{t}}~ \varepsilon_\theta \right) \\
    &+ \sqrt{1 - \alpha_{t+1}} \left(\sqrt{\alpha_{t}}~ \varepsilon_\theta + \sqrt{1 - \alpha_{t}}~ z_{t} \right).
\end{aligned}
\end{equation}

\noindent\textbf{Video Generation Model.}
In this paper, we propose a visual-textual video generation model based on LDMs (Latent Difference Models) to integrate source video, visual and textual clues to achieve generic video editing. The training process of our video generation model is depicted in Fig.~\ref{fig:train-process}.
Following \cite{makeavideo,zhou2022magicvideo,wu2022tune}, we initialize our model with text-to-image LDMs and we modify the U-Net to capture and exploit temporal correlations via temporal convolution, temporal attention, and spatio-temporal attention~\cite{wu2022tune}.
Our video generation model takes both textual and visual clues as conditioning inputs. To obtain the textual clue, we employ BLIP-2 \cite{blip2} to extract the video caption. The caption is then transformed into a text embedding using the CLIP Text Model \cite{clip}, which serves as the key and value in the cross-attention mechanism within the U-Net.
To ensure the ability to incorporate image embedding, we randomly select one frame from the video as the reference frame at each iteration. The reference frame is encoded using CLIP Vision Model~\cite{clip} and added to residual blocks in U-Net. Note that text and image embeddings are applied to all the blocks in the U-Net.

\subsection{Mask-Guided Fusion}
\label{sec:mask-denoise}
\noindent\textbf{Source Video Masks.}
Since the visual clues (CLIP image embedding) are added directly to the intermediate features in the residual blocks instead of operations with spatial information (\eg, cross-attention), the model cannot precisely control the location of the reference object. Therefore, we leverage the masks of the protagonist in the source video to separate foreground and background for spatial control.
Specifically, we first extract the protagonist mask in the first frame with Segment Anything~\cite{sam}. Then XMem~\cite{cheng2022xmem} is adopted to track the mask across the whole video.

\noindent\textbf{Overall Fusion Process.}
During the inference stage, the video generation model employs the DDIM inversed latent code as a starting point to denoise the code using the target textual and visual clues, as indicated by Eq.~\ref{eq:ddim}.
Since the CLIP Vision Model~\cite{clip} encodes the whole reference image, the background in the reference image is inevitably introduced into the generated result. To address this, we mask out the desired reference part via Grounded-SAM~\cite{sam,groundingdino}. 
However, when attempting to change the background of the source video, \eg \textit{``in the desert''} in Fig.~\ref{fig:framework}, the editing process may not yield satisfactory results due to the masked reference image.
The reason behind this is that the reference image embedding, which is directly added to the features, exerts a more dominant influence compared to the textual clues employed by the cross-attention mechanism.
Therefore, we incorporate the DALL-E~2 Prior~\cite{unclip}, capable of converting text embeddings into image embeddings, to enhance the representation of textual clues. The prior embedding and reference image embedding are fused together with the source masks.

Despite splitting the protagonist and background with the source masks, the protagonist and background may not be consistent across the video as the spatial information introduced by the source masks is quite coarse. 
Instead, self-attention maps can serve as good guidance for spatial location in each frame, and the spatial locations across the source video are consistent. Therefore, we can seek the background self-attention maps from the source video and use these maps to extract information from the target background. The attention fusion technique can improve the representation of the target background while keeping a similar temporal consistency as the source video. As for the protagonist part, we additionally leverage ControlNet~\cite{zhang2023adding} to provide more precise spatial control.
Feature fusion, attention fusion, and ControlNet are introduced as follows.

\noindent\textbf{Feature Fusion.}
Given the latent feature $z_t^F \in \mathbb{R}^{F\times C\times H\times W}$ in residual block, the image embedding $\mathcal{I} \in \mathbb{R}^{1\times C}$ is added to all frames and spatial location. 
We leverage the video protagonist masks $M \in \mathbb{R}^{F\times H\times W}$ to fuse the reference image embedding $\mathcal{I}_R \in \mathbb{R}^{1\times C}$ and prior converted text embedding $\mathcal{I}_P \in \mathbb{R}^{1\times C}$:
\begin{equation}
    z_t^F = z_t^F + M \cdot \mathcal{I}_R + (1-M)\cdot \mathcal{I}_P.
\end{equation}

In addition, since $\mathcal{I}_R$ represent the information of the protagonist, we introduce a timestamp parameter $\tau_F$ to control the ending step of the feature fusion operation:
\begin{equation}
    \label{eq:feature-fusion}
    z_t^F =  
    \begin{cases}
    z_t^F + M \cdot \mathcal{I}_R + (1-M)\cdot \mathcal{I}_P & \text { if } t < \tau_F \\ 
    z_t^F + \mathcal{I}_R & \text { otherwise. }
    \end{cases}
\end{equation}

Since the spatial location is determined in the early steps of the denoising process while details are refined in the latter stages, introducing $\tau_F$ can lead the model to contain more detailed information about the reference image.

\noindent\textbf{Attention Fusion.}
For the self-attention latent feature $z_t^A \in \mathbb{R}^{F\times C\times H\times W}$, frame-wise self-attention maps $A_t \in \mathbb{R}^{F\times HW\times HW}$ are calculated from the query and key projected features $Q(z_t^A)$ and $K(z_t^A)$. We leverage the video protagonist masks $M \in \mathbb{R}^{F\times H\times W}$ to fuse these self-attention maps with the source video self-attention maps $A_t^S$ obtained during DDIM inversion:
\begin{equation}
A_t = M \cdot A_t + (1-M) \cdot A_t^S.
\end{equation}

Similarly, a timestamp parameter $\tau_A$ is introduced to control the ending step of the attention fusion operation:
\begin{equation}
    \label{eq:attn-fusion}
    A_t =  
    \begin{cases}
    A_t = M \cdot A_t + (1-M) \cdot A_t^S & \text { if } t < \tau_A \\ 
    A_t & \text { otherwise. }
    \end{cases}
\end{equation}

\noindent\textbf{ControlNet.}
Source masks are coarse spatial constraint that applies the same information within them. 
Incorporating visual clues is difficult and such coarse constraint cannot lead to satisfactory results in most cases. 
Therefore, we additionally use a ControlNet~\cite{zhang2023adding} to enhance the spatial requirements within the mask. Specifically, the ControlNet branch takes one type of condition (pose or depth in this paper) together with the CLIP image embedding of the reference image. Then the ControlNet features are masked out by the source video masks and added to intermediate features in the diffusion U-Net.
Equipped with feature fusion, attention fusion and ControlNet, \ours can generate consistent video with both visual and textual clues.

\subsection{Applications of \ours}
\label{sec:applications}

We divide a video into background and protagonist, and \ours can edit either or both of them. Fig.~\ref{fig:teaser} illustrates three applications supported by our method.
In this section, we denote the reference image embedding $\mathcal{I}_R$ and the prior converted text embedding $\mathcal{I}_P$ in Eq.~\ref{eq:feature-fusion} as protagonist embedding and background embedding.

\noindent\textbf{Protagonist Editing:} 
In this application (2nd row in Fig.\ref{fig:teaser}), the background of the edited video should remain the same as the source video. Thus, instead of employing DALL-E~2 Prior~\cite{unclip} to convert the text into background embedding, we utilize the CLIP Vision Model~\cite{clip} to encode each frame of the source video as the background embedding ($\mathcal{I}_P$), while the image embedding from the reference image serves as the protagonist embedding ($\mathcal{I}_R$).

\noindent\textbf{Background Editing:}
To edit the background while preserving the protagonist (3rd row in Fig.~\ref{fig:teaser}), we encode the frames of the source videos as the protagonist embedding $\mathcal{I}_R$, and the text describing the background is converted to background embedding $\mathcal{I}_P$ by DALL-E~2 Prior~\cite{unclip}.

\noindent\textbf{Text-to-Video Editing with Protagonist:} 
By leveraging a reference image and text, both the background and protagonist can be edited simultaneously (4th row in Fig.\ref{fig:teaser}). In this application, the reference image is encoded as the protagonist embedding ($\mathcal{I}_R$) using the CLIP Vision Model~\cite{clip}, and the text is transformed into background embedding ($\mathcal{I}_P$) using DALL-E~2 Prior~\cite{unclip}.

\begin{figure*}[t]
    \centering
    \makebox[\textwidth][c]{\includegraphics[width=.99\textwidth]{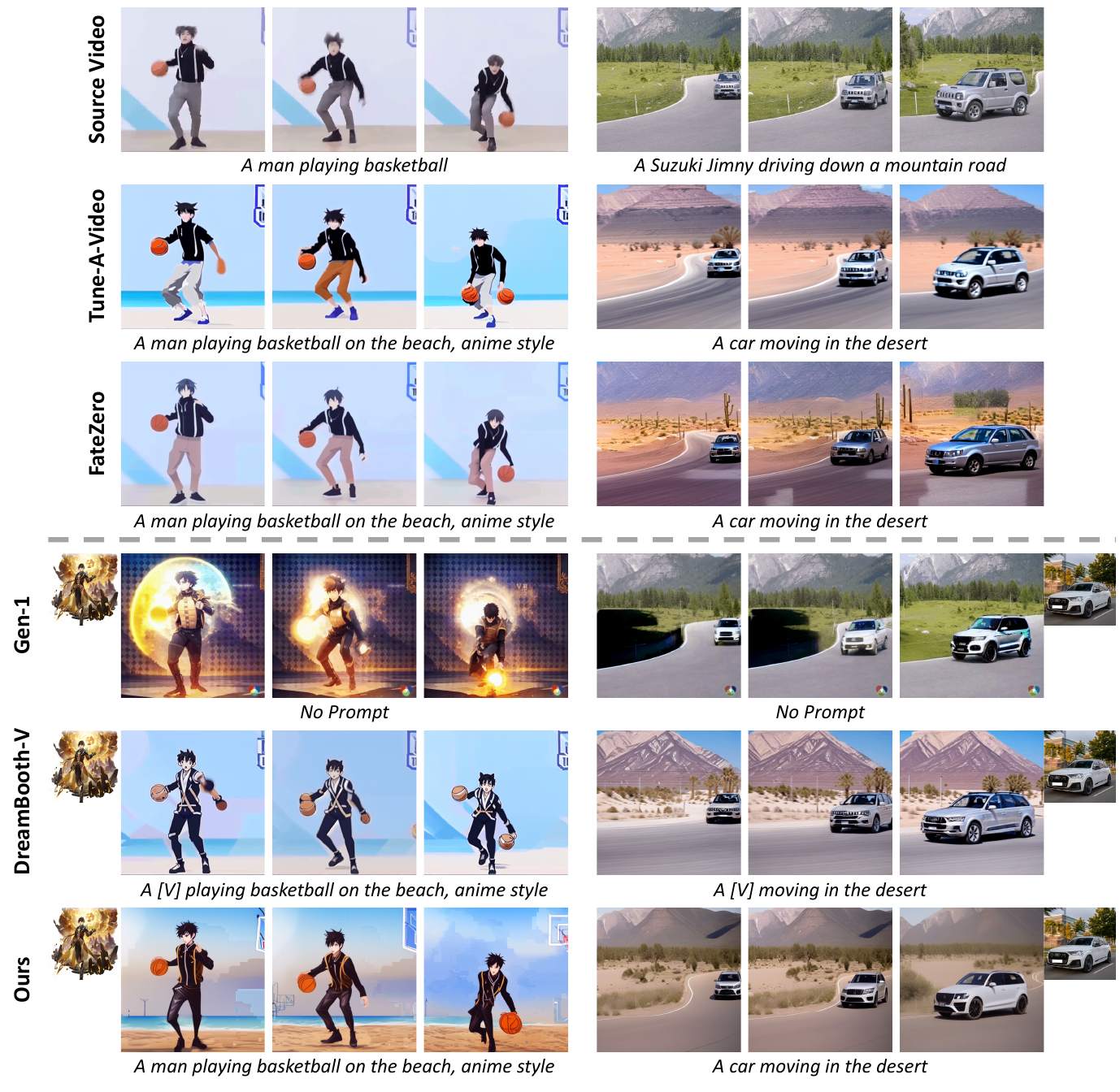}}
    \vspace{-.25in}
    \caption{Comparison with other video generation and editing methods. \ours can change the protagonist with the visual clue while editing the background with the textual clue.}
    \label{fig:sota}
\end{figure*}

\section{Experiments}

\subsection{Implementation Details}
Our video generation model is based on text-to-image (T2I) latent diffusion models with image embedding~\cite{stablediffusion} (Stable UnCLIP).
After initializing with the T2I LDM and inserting the temporal modules into the model, the video generation model is fine-tuned on 8 frames from a video at a resolution of 768 $\times$ 768. The model is trained for 200 steps with a learning rate of $3\times 10^{-5}$.
During inference, we use DDIM sampler~\cite{song2020denoising} with classifier-free guidance~\cite{ho2022classifier} for 20 steps.
It takes about 10 minutes to train a single video and 30 seconds for inference.

\begin{table*}[t]
\centering
\caption{Quantitative evaluation. \ours achieves comparable model evaluation with DreamBooth-V while overwhelming preference in the user study.}
\begin{tabular}{l|cc|ccc}
\toprule
& \multicolumn{2}{c|}{Model Evaluation} & \multicolumn{3}{c}{User Study} \\
& CLIP-T $\uparrow$ & DINO $\uparrow$ & Quality $\uparrow$ & Subject $\uparrow$ & Prompt $\uparrow$ \\
\midrule
\ours &  \textbf{0.337}  & 0.485 & \textbf{71.6\%} & \textbf{67.2\%} & \textbf{69.8\%}\\
DreamBooth-V~\cite{ruiz2022dreambooth,wu2022tune} & 0.301 & \textbf{0.509} & 21.0\% & 21.2\% & 12.4\% \\
Both Equally & --- & --- & 7.4\% & 11.6\% & 17.8\% \\
\bottomrule
\end{tabular}

\label{tab:quantitative-eval}
\end{table*}

\begin{figure*}[t]
    \centering
    \includegraphics[width=.9\textwidth]{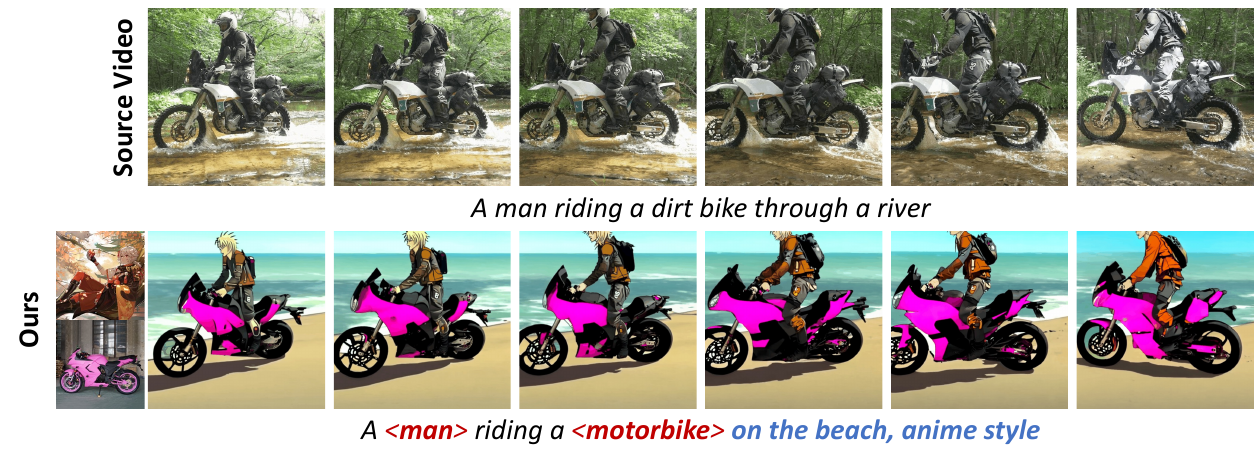}
    \vspace{-.15in}
    \caption{Example of editing two protagonists in one video. 
    }
    \vspace{-.1in}
    \label{fig:double-edit}
\end{figure*}

\subsection{Comparison with Baselines}

\textbf{Baselines.}
Since there are no previous methods conducting generic video editing, we compare \ours with text-based editing~\cite{wu2022tune, qi2023fatezero}, video variation~\cite{gen1}, and personalized~\cite{ruiz2022dreambooth} methods:
(1) \textit{Tune-A-Video}~\cite{wu2022tune} fine-tunes an inflated LDM on a single video to generate related content.
(2) \textit{FateZero}~\cite{qi2023fatezero} edits an existing video by controlling the cross-attention maps.
(3) \textit{Gen-1}~\cite{gen1} uses reference images or text to change the source video. Since Gen-1 can only use one condition (image or text), we use the reference image without text input.
(4) \textit{DreamBooth-V}~\cite{ruiz2022dreambooth,wu2022tune} is a baseline with a similar application as ours, combining DreamBooth~\cite{ruiz2022dreambooth} and Tune-A-Video~\cite{wu2022tune}. Specifically, we first train a personalized DreamBooth model with one reference image and the corresponding text token [V] and then use this personalized model as the initialization for single video fine-tuning~\cite{wu2022tune}. During inference, text token [V] can be used to generate content with the protagonist.

\noindent\textbf{Qualitative Evaluation.}
We compare the baselines on two videos in Fig~\ref{fig:sota}.
\textbf{First}, text-based editing methods~\cite{wu2022tune,qi2023fatezero} cannot integrate the visual clues, limiting their capability to generate indescribable protagonists.
\textbf{Second}, Gen-1~\cite{gen1} cannot use both visual and textual clues simultaneously for generic editing. Moreover, Gen-1~\cite{gen1} incorporates the background information from the reference image in the generated results.
\textbf{Third}, DreamBooth-V~\cite{ruiz2022dreambooth,wu2022tune} can perform generic video editing as \ours. However, one reference image is not enough to learn a good DreamBooth model for complex protagonists (\eg, the man in the left video) and thus DreamBooth-V struggles to generate realistic videos. Furthermore, even though the personalized model is well-trained, it requires a separate model for each protagonist with a careful selection of training configurations.
In addition, DreamBooth-V is limited to editing one protagonist in a video while \ours demonstrates the ability to edit multiple protagonists in Fig.~\ref{fig:double-edit}.

\begin{figure*}[t]
    \centering
    \makebox[\textwidth][c]{\includegraphics[width=1\textwidth]{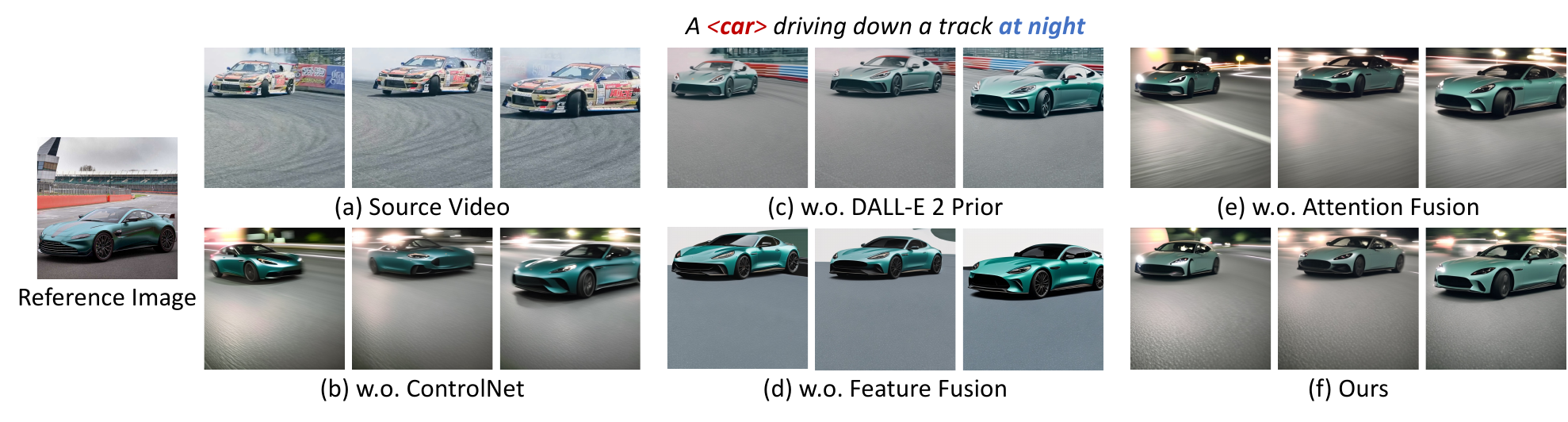}}
    \vspace{-.25in}
    \caption{Qualitative ablation studies. Removing the proposed components impairs the subject fidelity, prompt fidelity and video quality. }
    \label{fig:ablation}
\end{figure*}

\noindent\textbf{Quantitative Evaluation.}
We conduct a quantitative comparison with DreamBooth-V~\cite{ruiz2022dreambooth,wu2022tune} on generic video editing. Following~\cite{ruiz2022dreambooth}, prompt fidelity is measured by the average cosine similarity between the prompt and each video frame ViT-B/32 CLIP~\cite{clip} embeddings, denoted as CLIP-T. Subject fidelity is measured by the average cosine similarity between the ViT-S/16 DINO~\cite{dino} embeddings of each generated video frame and the reference image, denoted as DINO. 
In addition, we conduct a user study to compare the two methods in terms of video quality, subject fidelity, and prompt fidelity. We asked 20 users to answer questionnaires of 25 comparative questions. 
Each comparative question shows the reference image, text prompt, and videos generated by the two methods in random order. 
For video quality, users are asked to answer the question: ``Which of the two videos is more natural and realistic?'', and we include a ``Cannot Determine/Both Equally'' option.
Similarly, we ask ``Which of the two videos better reproduces the identity (e.g. item type and details) of the reference image?'' and ``Which of the two videos is better described by the text? (No need to take the reference image into account)'' for subject fidelity and prompt fidelity, respectively.
As shown in Tab.~\ref{tab:quantitative-eval}, \ours achieves a better prompt fidelity evaluated by CLIP while DreamBooth-V obtains a higher DINO score. However, we conjecture that the DINO score does not fully reflect the generated quality, since DreamBooth-V tends to overfit to the reference image. The overfitting leads to less realistic generated results but a higher DINO score (refer to qualitative comparison in the supplementary material).
Regarding the user study, we find an overwhelming preference for \ours in terms of video quality, subject fidelity, and prompt fidelity, demonstrating the superiority of our method.

\subsection{Ablation Studies}

\begin{table}[t]
\centering
\caption{Quantitative ablation studies.}
\vspace{-.1in}
\begin{tabular}{l|cc}
\toprule
 & CLIP-T $\uparrow$ & DINO $\uparrow$ \\
\midrule
w.o ControlNet~\cite{zhang2023adding} &  0.324 & 0.482 \\
w.o. DALL-E~2 Prior~\cite{unclip} & 0.297 & \textbf{0.490} \\
w.o. Feature Fusion & 0.293 & 0.472 \\
w.o. Attention Fusion & 0.335 & 0.479 \\
\midrule
\ours & \textbf{0.337} & 0.485 \\
\bottomrule
\end{tabular}
\label{tab:ablation}
\end{table}

To verify the effectiveness of each component in \ours, we conduct ablation studies on the ControlNet~\cite{zhang2023adding}, DALL-E 2 Prior~\cite{unclip}, feature fusion and attention fusion both quantitatively (Tab.~\ref{tab:ablation}) and qualitatively (Fig.~\ref{fig:ablation}).

\noindent\textbf{ControlNet}. Removing ControlNet can preserve the protagonist information but the motion is not consistent. In the first row of Tab.~\ref{tab:ablation}, removing ControlNet slightly influences the subject fidelity (DINO score) but impairs prompt fidelity (CLIP-T score).
In addition, due to lacking precise spatial location for the protagonist, the orientation of the car is wrong in the first several frames (Fig.~\ref{fig:ablation}(b)).

\noindent\textbf{DALL-E 2 Prior.} Deactivating DALL-E 2 Prior results in the text embedding being solely used in cross-attention. Consequently, the textual clue becomes weaker compared to the visual clue, leading to generating results that fail to align with the text description. Instead, since the CLIP image embedding for the background is zero, it would not influence the representation of the reference image. Therefore, the image similarity is slightly increased.

\noindent\textbf{Feature Fusion.}
Feature fusion is crucial for maintaining the visual information, resulting in a 0.013 improvement in DINO score. Furthermore, removing feature fusion equals to integrating the CLIP image embedding of the reference image to the whole frames, and the DALL-E 2 Prior embedding is also excluded.  Thus, the prompt fidelity is poor when deactivating feature fusion, which is also demonstrated in Fig.~\ref{fig:ablation}(d).

\noindent\textbf{Attention Fusion.}
Attention fusion aims to improve the temporal consistency by providing spatial information for the background. Comparing Fig.~\ref{fig:ablation}(e) and Fig.~\ref{fig:ablation}(f), attention fusion can address the inconsistent lane markings.
With the integration of DALL-E 2 Prior and mask-guided fusion, \ours showcases remarkable performance in both quantitative and qualitative assessments.

\section{Conclusion}
This paper introduces \ours, the first end-to-end framework for generic video editing using textual and visual clues. To edit the protagonist and background with these clues, we design a visual-textual-based video generation model coupled with a mask-guided fusion method to integrate diverse information sources.
The fusion approach effectively employs the masks, control signals and attention maps from the source video to provide precise spatial locations for editing both protagonist and background. 
By combining the video generation model with this fusion approach, \ours empowers versatile and powerful generic video editing applications, including background editing, protagonist editing, and text-to-video editing with the protagonist.

\noindent\textbf{Limitations \& Social Impacts.} 
One limitation of our approach is that representing visual clues with CLIP image embedding may not be optimal. It may struggle to encompass the full range of possible variations in a subject.
Additionally, the effectiveness of visual representation varies depending on the subject; for instance, our model demonstrates superior performance with cars compared to humans. Regarding social impacts, while personalized image generation offers numerous benefits, it simultaneously heightens the risk of misuse. Therefore, it is better to introduce a safety check for the source video and reference image to alleviate the possible negative impacts.

{
    \small
    \bibliographystyle{ieeenat_fullname}
    \bibliography{reference}
}

\end{document}